\pdfoutput=1
\documentclass[11pt]{article}

\PassOptionsToPackage{table,dvipsnames}{xcolor}
\usepackage[final]{acl}

\usepackage{algorithm}
\usepackage{algpseudocode}
\usepackage{amsmath}
\usepackage{booktabs}
\usepackage{multirow}
\usepackage{multicol}
\usepackage{caption}
\usepackage{subcaption}
\usepackage{tcolorbox} 
\usepackage{listings}
\usepackage{times}
\usepackage{latexsym}
\usepackage{subcaption}
\usepackage{microtype}
\usepackage{tabularx}
\usepackage{inconsolata}
\usepackage{graphicx}
\usepackage{diagbox}
\usepackage{array}

\definecolor{lightgray}{gray}{0.93}

\usepackage[T1]{fontenc}
\usepackage[utf8]{inputenc}

\usepackage{microtype}

\usepackage{inconsolata}

\usepackage{graphicx}

\title{Debating the Unspoken: Role-Anchored Multi-Agent Reasoning for Half-Truth Detection}
\author{Yixuan Tang \quad 
        Yirui Zhang \quad 
        Hang Feng \quad  
        Anthony Kum Hoe Tung  \\
School of Computing, National University of Singapore \\
\texttt{\{yixuan, yirui\_z, fenghang, atung\}@comp.nus.edu.sg}
}

\tcbuselibrary{breakable,listings}
\begin{document}
\maketitle
\begin{abstract}

Half-truths, claims that are factually correct yet misleading due to omitted context, remain a blind spot for fact verification systems focused on explicit falsehoods. Addressing such omission-based manipulation requires reasoning not only about what is said, but also about what is left unsaid. We propose \textsc{RADAR}, a role-anchored multi-agent debate framework for omission-aware fact verification under realistic, noisy retrieval. RADAR assigns complementary roles to a \emph{Politician} and a \emph{Scientist}, who reason adversarially over shared retrieved evidence, moderated by a neutral \emph{Judge}. A dual-threshold early termination controller adaptively decides when sufficient reasoning has been reached to issue a verdict. Experiments show that RADAR consistently outperforms strong single- and multi-agent baselines across datasets and backbones, improving omission detection accuracy while reducing reasoning cost. These results demonstrate that role-anchored, retrieval-grounded debate with adaptive control is an effective and scalable framework for uncovering missing context in fact verification. The code is available at \url{https://github.com/tangyixuan/RADAR}.

\end{abstract}
 % Unlike prior debate systems, RADAR operates on noisy retrieved evidence and dynamically balances reasoning depth with efficiency.

\section{Introduction}

\begin{figure}[t]
    \centering
    \includegraphics[width=0.96 \columnwidth]{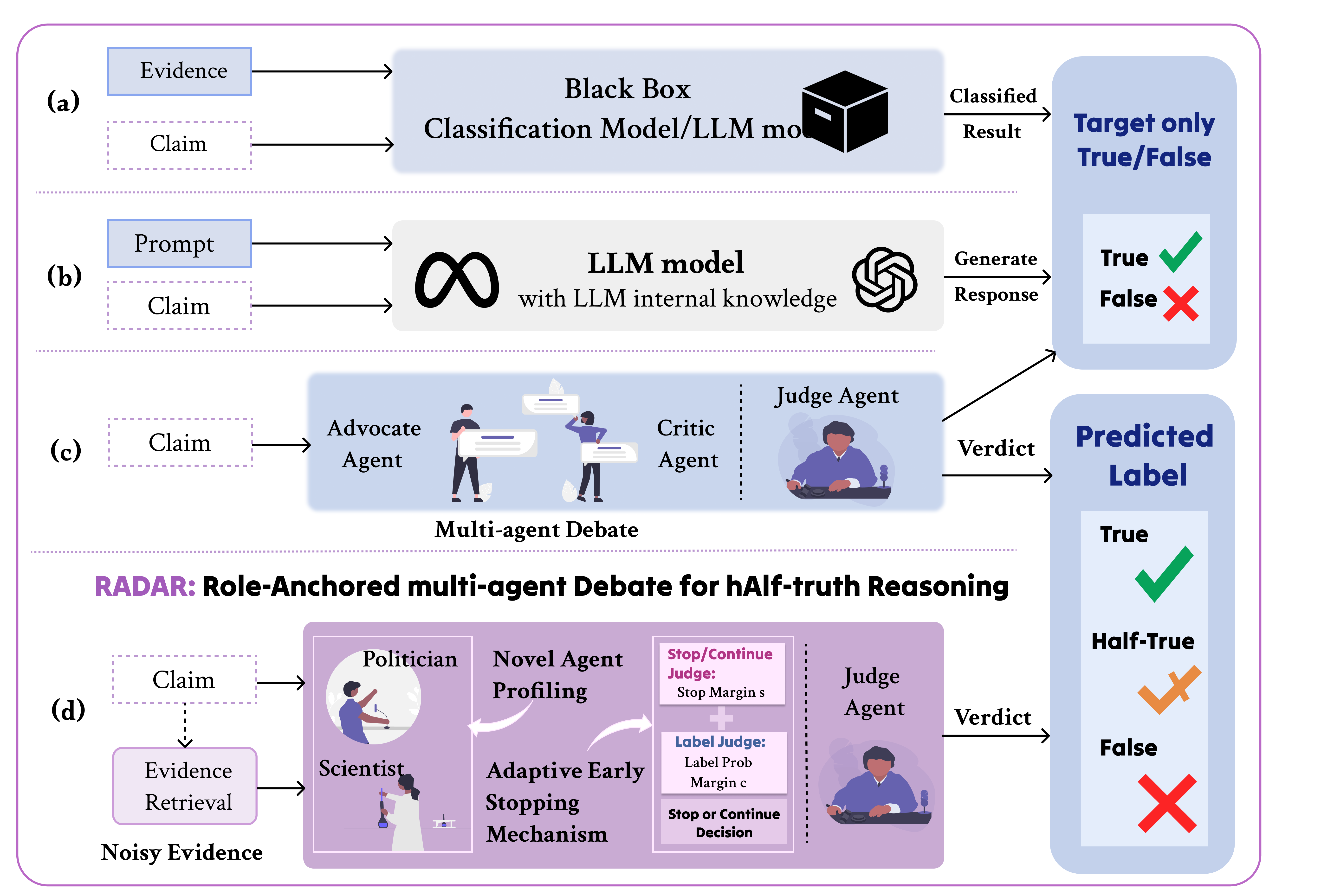}
    \caption{Overview of fact verification paradigms: 
    (a) encoder-based classifiers over claim–evidence pairs, 
    (b) prompt-guided LLMs, 
    (c) multi-agent debate with fixed \textit{pro/con} roles and internal knowledge, 
    and (d) our proposed \textsc{RADAR}, which performs omission-aware reasoning through evidence-grounded role debate.}
    \label{fig:types}
\end{figure}

The rapid spread of online misinformation has intensified the demand for scalable and reliable fact verification systems~\cite{Sanchaita2024to,Andrew2020deception}. While large language models (LLMs) have made notable progress in detecting explicit falsehoods, a subtler and more pervasive form of manipulation remains difficult to identify: the \emph{half-truth}. Half-truths are claims that are factually correct in isolation but misleading due to omitted context. For instance, stating that a politician reduced national debt by 15\% may be accurate, yet misleading if it conceals that the reduction followed a prior 20\% increase during the same term. Such selective framing distorts interpretation without fabricating facts, posing unique challenges for both human and automated verifiers~\cite{Sandeep2023beware}.

% The rapid spread of online misinformation has intensified the demand for scalable and reliable fact verification systems~\cite{openai2024gpt4technicalreport,Sanchaita2024to,Andrew2020deception}. While large language models (LLMs) have made notable progress in detecting explicit falsehoods, a subtler and more pervasive form of manipulation remains difficult to identify: the \emph{half-truth}. Half-truths are claims that are factually correct in isolation but misleading due to omitted context. For instance, stating that a politician reduced national debt by 15\% may be accurate, yet misleading if it conceals that the reduction followed a prior 20\% increase during the same term.

Detecting half-truths differs fundamentally from standard fact verification. Rather than identifying explicit contradictions, omission-aware verification must determine whether the available evidence is insufficient to justify a claim’s implied conclusion, requiring models to reason over latent intermediate assumptions rather than surface support signals. This challenge is exacerbated under realistic retrieval settings, where evidence is noisy and incomplete. Even when retrieved passages appear supportive, they may omit critical background information, causing systems to misclassify omission-based claims as true.

Most existing fact verification pipelines are not designed for this setting. As shown in Figure~\ref{fig:types}, single-agent approaches perform one-pass reasoning over retrieved evidence and are prone to error propagation when key context is missing. Multi-agent debate (MAD) frameworks~\cite{DBLP:conf/sigir/LiuLZC025,DBLP:conf/emnlp/HanZT25} introduce reasoning diversity through argument exchange, but their fixed \textit{pro/con} roles and focus on explicit contradiction make them ill-suited for omission reasoning, where the core issue is missing context rather than opposing claims. Recent work \textsc{TRACER}~\cite{DBLP:conf/emnlp/TRACER} has begun to explicitly study omission-aware verification, but assumes access to gold evidence and operates as a single-agent pipeline without adaptive reasoning, limiting its applicability under realistic retrieval conditions.

In this paper, we propose \textsc{RADAR} (\textbf{R}ole-\textbf{A}nchored multi-agent \textbf{D}ebate for h\textbf{A}lf-t\textbf{R}uth reasoning), a framework for omission-aware fact verification under incomplete retrieval. Unlike conventional debate systems with fixed \textit{pro/con} stances or internal knowledge reliance, \textsc{RADAR} formulates verification as a structured debate among agents with realistic expertise roles, all constrained to reason over retrieved evidence. A \emph{Politician} agent assesses the claim through public-interest and policy-communication considerations, while a \emph{Scientist} agent probes for missing, weak, or selectively presented context. Neither role is assigned a predefined \textit{pro/con} stance. Their interaction is moderated by a neutral \emph{Judge}, which synthesizes the debate and issues the final verdict.

To enhance efficiency, RADAR incorporates a dual-threshold early termination controller that monitors both the judge’s stop margin and verdict confidence. The debate concludes automatically once sufficient information has been exchanged, preventing redundant interactions while preserving reasoning depth. Through this adaptive mechanism, RADAR balances interpretability, accuracy, and computational cost, making omission reasoning more scalable for real-world verification.

% In this paper, we propose \textbf{RADAR} (\textbf{R}ole-\textbf{A}nchored multi-agent \textbf{D}ebate for h\textbf{A}lf-t\textbf{R}uth reasoning), a framework for omission-aware fact verification under incomplete retrieval. Unlike conventional debate systems with fixed \textit{pro/con} stances, \textsc{RADAR} formulates verification as a structured debate among agents with realistic expertise roles, all constrained to reason over retrieved evidence. A \emph{Politician} agent constructs a coherent narrative supporting the claim, while a \emph{Scientist} agent probes for missing or selectively presented context. Their interaction is moderated by a neutral \emph{Judge}, which synthesizes the debate and issues the final verdict. A dual-threshold early termination controller further monitors the judge's stop margin and verdict confidence, concluding debate automatically once sufficient reasoning has been reached to balance accuracy and computational cost.

Our contributions are as follows:
\begin{itemize}
    % \item We address omission-based half-truth detection under a realistic retrieval setting, highlighting the challenges of identifying missing context from incomplete evidence.
    % \item We propose RADAR, a role-anchored and retrieval-grounded multi-agent reasoning framework that surfaces missing context through complementary debate, with adaptive early stopping for efficiency.
    \item We address the underexplored challenge of half-truth detection under realistic noisy retrieval, where gold evidence is unavailable.
    \item We propose RADAR, a role-anchored, retrieval-grounded multi-agent framework that surfaces omission-based half-truths through complementary expertise debate, with adaptive early stopping for efficiency.
    \item We demonstrate across datasets and backbone LLMs that RADAR consistently improves both accuracy and efficiency over strong single-agent and multi-agent baselines.
\end{itemize}

\section{Related Work}

\subsection{Fact Verification and Half-Truths}

Fact verification is commonly formulated as determining whether a claim is supported or refuted given retrieved evidence~\cite{thorne2018fever,alam2021fighting,zhang2023hiss,wu2025system}, with numerous systems exploring retrieval, reasoning, and explanation~\cite{chen2022gerearr,sriram2024constrastive,yue2024retrieval,wang2025maferw}. \textsc{Climinator} \cite{DBLP:journals/corr/abs-2401-12566} studies LLM-based fact verification in the climate domain. \textsc{ClaimDecomp}~\cite{implicit-question} decomposes claims into sub-questions, while \textsc{FIRE}~\cite{fire2023} uses iterative verify-or-search loops to reformulate queries upon failure. While effective for explicit misinformation, these approaches largely assume that misleading claims manifest as direct contradictions with evidence.

Half-truths violate this assumption. Claims may be factually correct with respect to retrieved evidence yet still be misleading due to omitted context. \textsc{TRACER}~\cite{DBLP:conf/emnlp/TRACER} is the first framework to explicitly model this phenomenon through the concept of \emph{critical hidden evidence}, but assumes access to gold evidence annotations and follows a single-agent, static reasoning pipeline. Our work addresses omission detection under noisy retrieval through structured multi-agent interaction.

\begin{figure*}[t]
\centering
\includegraphics[width=0.99\linewidth]{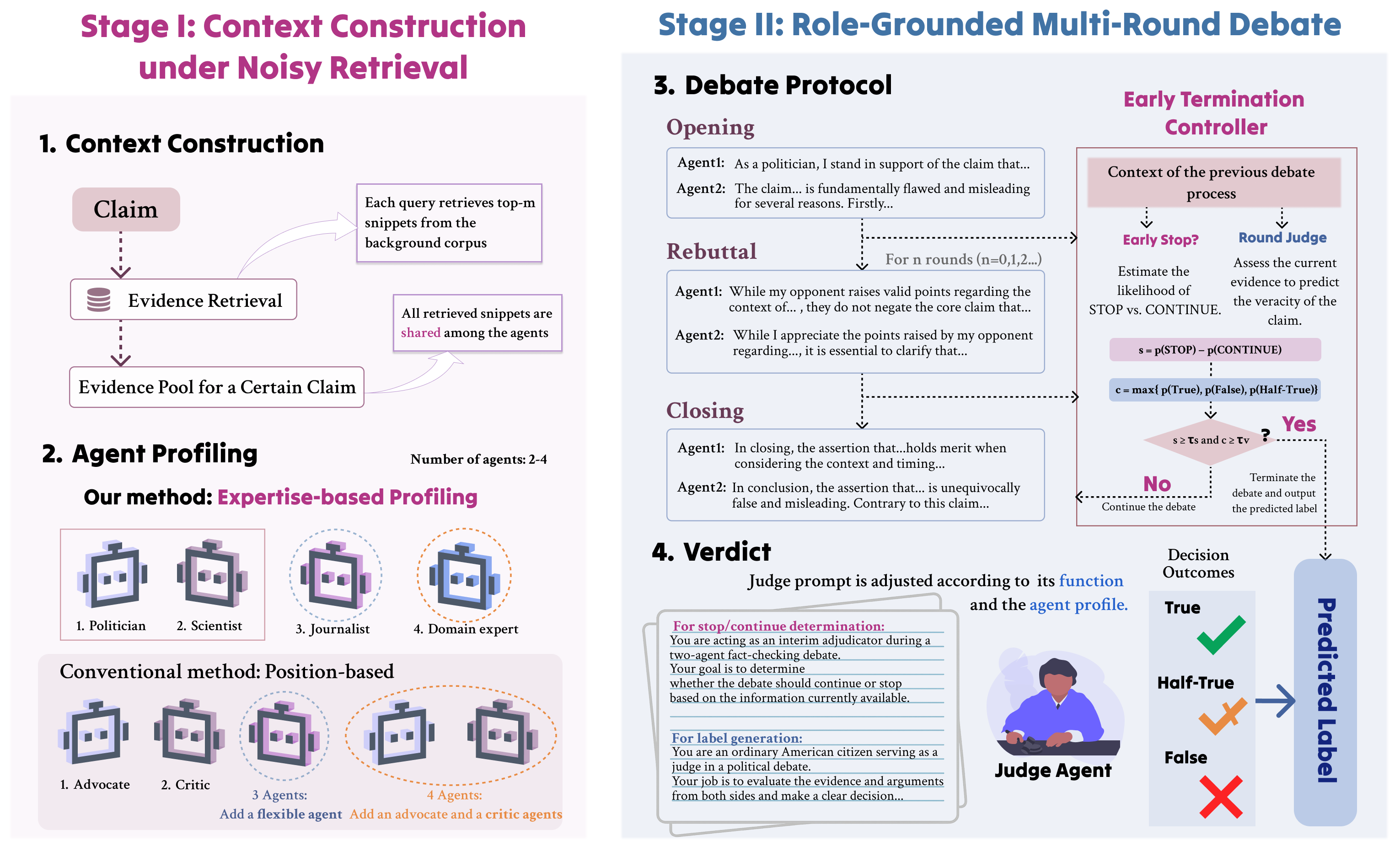}
\caption{Overview of the \textsc{RADAR} framework for omission-based half-truth detection. The system conducts structured multi-agent debate between expertise-grounded roles over retrieved evidence, equipped with an adaptive early termination controller to uncover missing yet critical context efficiently. The Judge is fixed and excluded from the reported number of debating agents.}
\label{fig:framework}
\end{figure*}

\subsection{Multi-Agent Debate for Reasoning}

Multi-agent debate (MAD) has recently emerged as a paradigm for enhancing reasoning and factuality in LLMs~\cite{DBLP:conf/icml/Du00TM24,wei2022chain,liang2023divergent,guo2024llmbased,tillmann2025,DBLP:conf/sigir/LiuLZC025}. Debate-based frameworks such as D2D~\cite{DBLP:conf/emnlp/HanZT25} and TruEDebate (TED)~\cite{DBLP:conf/sigir/LiuLZC025} employ multi-stage dialogue between a proponent, skeptic, and judge to reach veracity judgments for fake news. However, most MAD systems are designed for contradiction-oriented verification, where agents argue for or against a claim. Omission-based half-truths differ fundamentally: the key issue may be missing context rather than explicit conflict. RADAR therefore replaces polarity-based debate with complementary reasoning roles and grounds all interaction in retrieved evidence, enabling agents to surface what is absent rather than only dispute what is stated.

% \vspace{-5pt}

\subsection{Bias by Omission in News}

\citet{DBLP:conf/isiwi/EhrhardtSVH21} detect political slant by analyzing entity co-occurrence across outlets. \citet{DBLP:conf/jcdl/ZhukovaRHDG23} model bias by commission, omission, and source selection through the COSS framework. Other works explore rhetorical manipulation and propaganda framing~\cite{DBLP:conf/emnlp/MartinoYBPN19,DBLP:conf/asiaccs/AkhtarM0K24,DBLP:conf/acl/PiskorskiS0MN23} or examine linguistic cues distinguishing factual from deceptive reporting~\cite{DBLP:conf/emnlp/RashkinCJVC17}. These studies highlight omission as a key mechanism of bias. Our work formulates omission as a reasoning and verification task, aiming to detect when incomplete context leads to misleading claims.

\section{RADAR Framework}

\subsection{Framework Overview}

We present \textbf{RADAR} (\textit{Role-Anchored multi-agent Debate for hAlf-truth Reasoning}), a multi-agent framework for detecting omission-based half-truths under realistic retrieval conditions. Unlike prior systems that rely on single-pass classification or assume gold evidence, RADAR performs deliberative reasoning among role-specialized agents who exchange arguments grounded in retrieved evidence. As illustrated in Figure~\ref{fig:framework}, the framework operates in two stages: (1) \textit{Context Construction under Noisy Retrieval}, which builds a shared evidence pool without assuming completeness, and (2) \textit{Role-Grounded Multi-Round Debate}, where agents reason adversarially to identify missing context and assess whether the claim is misleading.

% Given an input claim $c$, RADAR first constructs an evidence pool $E$ by retrieving the top-ranked text segments for the input claim from a large background corpus. The retrieved set inevitably contains irrelevant or incomplete content, thereby simulating the uncertainty of real-world fact-checking scenarios. Three agents participate in the subsequent debate: a \textit{Politician}, a \textit{Scientist}, and a neutral \textit{Judge}. The  Politician agent is designed to construct a persuasive narrative from the available evidence, representing an advocacy-oriented reasoning style. In contrast, the Scientist agent critically examines the same evidence for missing, weak, or selectively presented support, representing an analytical reasoning style. Although the role names are inspired by political discourse, their functional behaviors are domain-agnostic and intended to capture complementary reasoning tendencies. The Judge moderates the discussion, monitors sufficiency, and issues the final verdict. This role design transforms verification from stance competition into omission-focused reasoning: one agent constructs the strongest evidence-backed narrative, while the other probes for missing or underrepresented context. Combined with shared evidence grounding and adaptive stopping, RADAR enables structured reasoning under incomplete retrieval.

Given an input claim $c$, RADAR constructs an evidence pool $E$ by retrieving top-ranked text segments from a background corpus. Two debating agents, a \textit{Politician} and a \textit{Scientist}, interact under a fixed neutral \textit{Judge}; their roles and reasoning styles are summarized in Table~\ref{tab:roles}. This design transforms verification from stance competition into omission-focused reasoning, with both debaters independently assessing the claim through complementary expertise rather than predefined \textit{pro/con} positions.

% This role design captures complementary reasoning tendencies, persuasive framing versus investigative scrutiny, revealing hidden context often overlooked by single-agent systems.

% The Politician seeks to explore and articulate the claim’s narrative, while the Scientist probes for missing or contradictory context. 

\begin{table}[t]
\centering
\small
\begin{tabular}{l|p{5.8cm}}
\hline
\textbf{Role} & \textbf{Objective and Reasoning Bias} \\
\hline
Politician & Assesses the claim through public-interest and policy-communication considerations; emphasizes narrative coherence. \\
Scientist & Examines retrieved evidence for missing or contradicting information; tends to challenge selective framing. \\
Judge & Evaluates debate sufficiency, coherence, and evidential coverage; decides when to stop and issues the final verdict. \\
\hline
\end{tabular}
\caption{Roles and reasoning types in \textsc{RADAR}.}
\label{tab:roles}
\end{table}

\subsection{Stage I: Context Construction under Noisy Retrieval}

Omission-based misinformation often remains convincing because relevant facts are selectively presented while critical context is absent. To evaluate claims under realistic conditions, RADAR first constructs a retrieval-grounded evidence pool that is informative yet imperfect. Rather than assuming complete supporting evidence, the system requires downstream reasoning to operate under uncertainty, where retrieved information may be partial, noisy, or misleading.

\paragraph{Evidence Retrieval.}
Given a claim $c$, RADAR retrieves the top-$m$ relevant snippets from a background corpus to form an evidence pool $E = \textsc{Retrieve}(c)$. Because retrieval is inherently imperfect, $E$ may contain irrelevant passages, fragmented facts, or only partially supportive evidence. This setting better reflects real-world fact verification, where systems must reason from incomplete retrieved context. It differs from prior omission-aware work such as \textsc{TRACER}~\cite{DBLP:conf/emnlp/TRACER}, which assumes access to gold evidence, and from debate frameworks such as \textsc{TED}~\cite{DBLP:conf/sigir/LiuLZC025} or \textsc{D2D}~\cite{DBLP:conf/emnlp/HanZT25}, which rely primarily on internal model knowledge.

\paragraph{Shared Grounding.}
The same evidence pool $E$ is provided to all agents as a common factual basis. During debate, agents are required to justify their arguments using retrieved evidence, which improves transparency and reduces unsupported claims. Since all agents observe identical but incomplete evidence, differences in their conclusions arise from reasoning rather than information asymmetry. % This shared-grounding design enables RADAR to uncover omitted context through adversarial scrutiny over the same retrieved facts.

\subsection{Stage II: Role-Grounded Multi-Round Debate}

Once the shared context is established, RADAR conducts a structured debate in which the Politician and Scientist iteratively exchange arguments while the Judge moderates and evaluates progress. % All agents operate on the same evidence pool $E$ and follow a fixed debate protocol.

\paragraph{Debate Protocol.}
Each debate instance begins with an \textit{opening round}, where both agents independently assess the claim from their role-specific perspectives and may support or challenge it. In the \textit{rebuttal round}, both agents are tasked with challenging the opponent’s argument by identifying potential flaws or weaknesses. Finally, in the \textit{closing round}, both summarize their stances, after which the Judge synthesizes the debate transcript and evidence to assign one of three labels: \textit{true}, \textit{half-true}, or \textit{false}. For AVeriTeC, NEI is retained as a fourth label.
% to decide whether the claim is fully supported, due to omitted context, or contradicted by the evidence.

\paragraph{Adaptive Early Stopping.}

To balance reasoning thoroughness and efficiency, RADAR introduces a dual-threshold \textit{early termination controller} that determines whether to continue or stop after each round. At the end of every round, the Judge model evaluates the accumulated dialogue and computes the \textit{stop margin} as 
\[
s = p(\text{STOP}) - p(\text{CONTINUE})
\]
where $p(\cdot)$ denotes the softmax-normalized probability of the corresponding decision token. The debate proceeds if $s < \tau_s$, and terminates otherwise. To prevent premature stopping on ambiguous cases (especially half-truths), we further require the model's veracity confidence
\[
c = \max_{y \in \mathcal{Y}} p(y)
\]
where $\mathcal{Y}=\{\text{True}, \text{False}, \text{Half-True}\}$. Early termination occurs only when $s \ge \tau_s$ \textit{and} $c \ge \tau_v$. Both thresholds are calibrated on a development set using cached decision scores from a single inference pass. This adaptive control balances reasoning depth with computational cost, allowing RADAR to maintain accuracy while shortening redundant interactions.

% This adaptive control preserves reasoning depth where needed while avoiding redundant interactions.

\paragraph{Structured Verdict and Interpretability.}
At the end of the debate, the Judge produces a structured output:

\vspace{0.3em}
\textbf{[REASON]}: \textit{Concise explanation citing evidence and debate exchanges.}

\textbf{[VERDICT]}: \textit{TRUE / FALSE / HALF-TRUE}

The inclusion of debate transcripts and evidence citations provides transparent reasoning paths that expose which contextual gaps or omissions influenced the final decision. This interpretability distinguishes RADAR from opaque end-to-end verifiers, enabling human fact-checkers to trace and validate each reasoning step.

By integrating retrieval-grounded context, expertise-based roles, and adaptive termination, RADAR transforms omission detection from static classification into a dynamic reasoning process. All prompt templates and implementation details are presented in Appendix~\ref{appendix:prompts}.

% The Politician–Scientist pairing introduces complementary cognitive biases, and the Judge ensures coherent, efficient, and interpretable deliberation. Together, these components enable RADAR to reason effectively under incomplete evidence, bridging the gap between conventional fact verification and omission-aware misinformation analysis.

\section{Experiments}

\subsection{Experimental Setup}

\paragraph{Dataset.}

We evaluate \textsc{RADAR} on the \textsc{PolitiFact-Hidden} benchmark~\cite{DBLP:conf/emnlp/TRACER}, which contains $\sim$15k political claims with gold evidence and three labels (\textit{true}, \textit{half-true}, \textit{false}), focusing on omission-based reasoning rather than explicit contradiction. The evidence corpus includes 41,952 documents, and we follow the original train/dev/test split (11,994/1,000/2,000). As \textsc{RADAR} is training-free, thresholds are tuned on the dev set and results are reported on the test set.

\begin{table}[t]
\centering
\small
\begin{tabular}{cccccc}
\toprule
\textbf{Model}  & \textbf{Accu.} & \textbf{F1}$_{mc}$ & \textbf{F1}$_{T}$ & \textbf{F1}$_{HT}$ & \textbf{F1}$_{F}$ \\ \midrule

\rowcolor{gray!10}
\multicolumn{6}{c}{\textit{Full Evidence Setting}} \\ \midrule

CoT  & 76.3 & 64.3 & 52.9 & 52.8 & 87.1 \\
CoT + RA  & 78.5 & 68.0 & 56.7 & 60.2 & 87.1 \\
HiSS  & 74.7 & 61.9 & 48.1 & 51.2 & 86.4 \\
HiSS + RA  & 77.8 & 65.7 & 49.1 & 61.5 & 86.5 \\ 
RADAR$_{\text{single}}$  & 68.7 & 62.4 & 57.6 & 51.1 & 78.6 \\
RADAR$_{\text{multi}}$  & 
\textbf{83.6}  & 
\textbf{70.8} & 
\textbf{59.3}  & 
\textbf{61.1} & 
\textbf{92.0}  \\ 
\midrule
\rowcolor{gray!10} 
\multicolumn{6}{c}{\textit{Retrieved Evidence Setting}} \\ \midrule

FIRE & 60.3 & 46.9 & 30.7 & 34.1 & 75.9 \\
D2D  & 63.0 & 50.9 & 36.3 & 39.7 & 76.5 \\
RADAR$_{\text{single}}$  & 58.4 & 51.0 & 41.2 & 41.5 & 70.2 \\
RADAR$_{\text{multi}}$  & 
\textbf{77.7}  & 
\textbf{63.3} & 
\textbf{45.6}  & 
\textbf{56.5}  & 
\textbf{87.6}  \\

\bottomrule
\end{tabular}
\caption{Main results under full and retrieved evidence settings using \texttt{GPT-4o-mini}. “+RA” denotes TRACER-enhanced baseline methods. \textsc{RADAR}$_{\text{single}}$ and \textsc{RADAR}$_{\text{multi}}$ share identical architectures and retrieval setups, differing only in whether multi-agent debate is used.}
\label{tab:main-results}
\end{table}

\begin{table*}[t]
\centering
\small
\begin{tabular}{ccccccc}
\toprule
\textbf{Model} & \textbf{Backbone} & \textbf{Accuracy} & \textbf{F1}$_{macro}$ & \textbf{F1}$_{true}$ & \textbf{F1}$_{half\text{-}true}$ & \textbf{F1}$_{false}$ \\ \midrule

% FIRE &  LLaMA3-8B-Instruct &50.4&43.4&30.9&39.7&59.6 \\
% D2D & LLaMA3-8B-Instruct &64.1&45.1&21.7&34.0&79.5 \\
FIRE &  \multirow{4}{*}{ GPT-4o mini} & 60.3 & 46.9 & 30.7 & 34.1 & 75.9 \\
D2D &  & 63.0 & 50.9 & 36.3 & 39.7 & 76.5 \\
RADAR$_{\text{single}}$ &  & 58.4 & 51.0 & 41.2 & 41.5 & 70.2 \\
RADAR$_{\text{multi}}$ &  & 
\textbf{77.7} {\textcolor{blue}{$\uparrow$ 19.3}} & 
\textbf{63.3} {\textcolor{blue}{$\uparrow$ 12.3}} & 
\textbf{45.6} {\textcolor{blue}{$\uparrow$ 4.4}} & 
\textbf{56.5} {\textcolor{blue}{$\uparrow$ 15.0}} & \textbf{87.6} {\textcolor{blue}{$\uparrow$ 17.4}} \\

 \midrule

FIRE &  \multirow{4}{*}{ LLaMA3-8B-Instruct} & 50.4 & 43.4 & 30.9 & 39.7 & 59.6 \\
D2D & & 64.1 & 45.1 & 21.7 & 34.1 & 79.5 \\
RADAR$_{\text{single}}$ &  & 57.3 & 52.0 & 44.4 & 44.8 & 66.6 \\
RADAR$_{\text{multi}}$ &  & 
\textbf{77.3} {\textcolor{blue}{$\uparrow$ 20.0}} & 
\textbf{63.2} {\textcolor{blue}{$\uparrow$ 11.2}} & 
\textbf{46.3} {\textcolor{blue}{$\uparrow$ 1.9}} & 
\textbf{56.7} {\textcolor{blue}{$\uparrow$ 11.9}} & 
\textbf{86.5} {\textcolor{blue}{$\uparrow$ 19.9}} \\

 \midrule
FIRE &  \multirow{4}{*}{ Qwen2.5-7B-Instruct} & 63.9 & 46.9 & 29.2 & 32.1 & 79.5 \\
D2D &  & 69.0 & 48.5 & 22.2 & 41.2 & 82.1 \\ 
RADAR$_{\text{single}}$ &  & 71.2 & 54.9 & 39.4 & 41.1 & 84.3 \\ % single 
RADAR$_{\text{multi}}$  &  &
\textbf{76.7} {\textcolor{blue}{$\uparrow$ 5.5}}&
\textbf{63.1} {\textcolor{blue}{$\uparrow$ 8.2}} &
\textbf{48.2} {\textcolor{blue}{$\uparrow$ 8.8}}&
\textbf{54.9} {\textcolor{blue}{$\uparrow$ 13.8}}&
\textbf{86.2} {\textcolor{blue}{$\uparrow$ 1.9}}\\ 
\bottomrule
\end{tabular}
\caption{Generalization results of strong baselines and \textsc{RADAR} variants across different backbone LLMs under the retrieved evidence setting.}
\label{tab:main-results-backbone}
\end{table*}

% \multicolumn{7}{c}{\textit{Retrieved Evidence Setting}} \\

\begin{table}[t]
\centering
\small
\setlength{\tabcolsep}{3pt}
\begin{tabular}{lcccccc}
\toprule
\textbf{Model} & \textbf{Acc.} & \textbf{F1}$_{mc}$ & \textbf{F1}$_{T}$ & \textbf{F1}$_{HT}$ & \textbf{F1}$_{F}$ & \textbf{F1}$_{\text{NEI}}$ \\ \midrule

FIRE & 26.8 & 21.1 & 11.9 & 14.7 & 45.6 & 12.1 \\
D2D & 63.8 & 46.9 & 55.3 & 16.9 & 80.4 & 35.1 \\
RADAR$_{\text{multi}}$ & \textbf{77.8} & \textbf{57.7} & \textbf{79.6} & \textbf{20.3} & \textbf{86.8} & \textbf{43.9} \\

\bottomrule
\end{tabular}
\caption{Four-class performance on AVeriTeC, where ``Conflicting Evidence/Cherry-picking'' is mapped to \textit{half-true} and NEI is retained as a separate class.}
\label{tab:more_data}
\end{table}

To assess generalization, we further evaluate \textsc{RADAR} on AVeriTeC~\cite{DBLP:conf/nips/SchlichtkrullG023}, a real-world claim verification benchmark that retrieves multi-step question-answer evidence from the open web. We map its ``Conflicting Evidence/Cherry-picking'' label to \textit{half-true} and retain \textit{not enough evidence} (NEI) as a fourth class.

\paragraph{Settings.}
% We evaluate under two conditions. In the \textit{full evidence} setting, models receive all gold evidence associated with each claim. In the \textit{retrieved evidence} setting, we simulate realistic verification by retrieving top-20 passages per claim from the corpus using a dense retriever (\texttt{bge-base-en-v1.5}). This introduces noise and incompleteness, testing whether RADAR’s debate mechanism can recover missing context when retrieval is imperfect. Unless otherwise specified, the same role prompts and debate protocol are used across all datasets and backbone models, with no dataset-specific prompt tuning or role adaptation.

% To isolate the contribution of multi-agent debate, we additionally include a single-agent variant that uses the same architecture, prompts, retrieval setup, and inference procedure, but without inter-agent interaction. We refer to this variant as RADAR$_{\text{single}}$, and denote the full multi-agent system as RADAR$_{\text{multi}}$.

We evaluate under two conditions. In the \textit{full evidence} setting, models receive all gold evidence per claim. In the \textit{retrieved evidence} setting, we retrieve top-20 passages per claim using a dense retriever (\texttt{bge-base-en-v1.5}), introducing noise and incompleteness to test whether RADAR's debate mechanism can recover missing context. Sensitivity to retrieval depths of $top-k \in \{1,3,5,10,20\}$, together with the corresponding gold-evidence recall, is detailed in Appendix~\ref{app:retrieval}. The same role prompts and debate protocol are used across all datasets and backbones, with no dataset-specific tuning.

To isolate the contribution of multi-agent debate, we include a single-agent variant (\textsc{RADAR}$_{\text{single}}$) with identical architecture, prompts, and retrieval but without inter-agent interaction. We denote the full system as \textsc{RADAR}$_{\text{multi}}$.

\paragraph{Baselines.}
We compare against representative fact verification and debate-based methods:

     \textbf{CoT}~\cite{wei2022chain}: Chain-of-thought prompting over provided evidence.
     
    \textbf{HiSS}~\cite{zhang2023hiss}: Hierarchical structured prompting with intermediate reasoning steps.
    
     \textbf{TRACER}~\cite{DBLP:conf/emnlp/TRACER}: An omission-aware verification framework that assumes access to complete gold evidence, including hidden context required for intent and assumption inference.
     
    \textbf{FIRE}~\cite{fire2023}: An iterative retrieve-then-verify framework alternating between retrieval and validation.
    
     \textbf{D2D}~\cite{DBLP:conf/emnlp/HanZT25}: A multi-agent debate model for news misinformation detection.

% We compare against representative fact verification and debate-based methods: \textbf{CoT}~\cite{wei2022chain}, chain-of-thought prompting over evidence; \textbf{HiSS}~\cite{zhang2023hiss}, hierarchical structured prompting with intermediate reasoning steps; \textbf{TRACER}~\cite{DBLP:conf/emnlp/TRACER}, an omission-aware framework assuming access to complete gold evidence; \textbf{FIRE}~\cite{fire2023}, an iterative retrieve-then-verify framework; and \textbf{D2D}~\cite{DBLP:conf/emnlp/HanZT25}, a multi-agent debate model with fixed \textit{pro/con} roles.

For all baselines, we follow the experimental settings recommended in their original papers to reflect their intended usage. For prompting-based methods such as CoT, we use a standard task instruction with step-by-step reasoning. For framework-specific baselines such as FIRE and D2D, we preserve their original prompting protocols. All shared components (backbone models, retrieval pipeline, evidence pool) are kept identical across methods to isolate the effect of reasoning frameworks. % Meanwhile, all shared components, including the backbone models, retrieval pipeline, and evidence pool, are kept identical across methods. This controlled setup helps isolate the effect of reasoning frameworks rather than differences in external resources or implementation details.

\paragraph{Evaluation Metrics.}
We report accuracy and macro-F1 as overall metrics, along with per-class F1 scores (\textit{true}, \textit{half-true}, \textit{false}) to analyze class-level behavior; for AVeriTeC, macro-F1 and per-class results additionally include NEI. Performance changes for \textsc{RADAR}$_{\text{multi}}$ are measured relative to its single-agent variant. Each debate proceeds for up to three rounds, with an adaptive early stopping mechanism. % To assess generality, we evalaute RADAR using three backbone LLMs: \texttt{GPT-4o-mini}, \texttt{LLaMA3-8B-Instruct} and \texttt{Qwen2.5-7B-Instruct}.

\subsection{Main Results}

Table~\ref{tab:main-results} presents the main comparison results.
\paragraph{Performance with Gold Evidence.}

When complete gold evidence is provided, most baselines achieve competitive performance, reflecting that omission-aware verification becomes easier when all relevant context is available. Nevertheless, \textsc{RADAR}$_{\text{multi}}$ attains the best overall performance, achieving 83.6 accuracy and 70.8 macro-F1, outperforming strong baselines such as CoT+RA and HiSS+RA by clear margins. Compared to \textsc{RADAR}$_{\text{single}}$, multi-agent debate yields substantial gains (+14.9 accuracy and +8.4 macro-F1), indicating that structured interaction between expertise-anchored agents improves reasoning consistency even when evidence is complete.

\paragraph{Performance with Retrieved Evidence.}
The retrieved-evidence setting is more realistic and challenging due to noisy and incomplete evidence. \textsc{RADAR}$_{\text{multi}}$ remains robust, achieving 77.7 accuracy and 63.3 macro-F1, substantially outperforming all strong baselines. Relative to its single-agent counterpart, multi-agent debate brings large improvements (+19.3 accuracy and +12.3 macro-F1), demonstrating that RADAR's structured interaction is particularly effective when critical evidence is missing or fragmented.

\paragraph{Per-Class Performance Analysis.}
Per-class results show that performance gains are concentrated in the \textit{half-true} and \textit{false} categories, where omission reasoning is most critical. Under retrieved evidence, \textsc{RADAR}$_{\text{multi}}$ improves \textit{half-true} F1  by 15 and \textit{false} F1 by 17.4, while also maintaining stronger performance on the \textit{true} class. The pronounced gap between \textsc{RADAR}$_{\text{single}}$ and \textsc{RADAR}$_{\text{multi}}$ on these classes suggests that debate helps scrutinize selectively framed narratives and surface omitted context beyond what single-pass reasoning can achieve. To further examine model behavior near class boundaries, especially the subtle distinction between \textsc{True} and \textsc{Half-True}, we complement the per-class F1 scores in Table~\ref{tab:main-results-backbone} with confusion matrices and class-wise precision/recall analyses in Appendix~\ref{app:boundary}. % These results show that RADAR achieves stronger discrimination on omission-sensitive cases rather than benefiting merely from boundary ambiguity.

\paragraph{Generalization Across Backbone LLMs.}
Table~\ref{tab:main-results-backbone} reports results across three backbone LLMs. \textsc{RADAR}$_{\text{multi}}$ consistently achieves strong performance across \texttt{GPT-4o-mini}, \texttt{LLaMA3-8B-Instruct}, and \texttt{Qwen2.5-7B-Instruct}. While absolute performance varies across backbones, the relative improvements over strong baselines and the single-agent variant remain stable, indicating that RADAR’s debate mechanism generalizes across architectures.

\paragraph{Generalization Across Fact Verification Datasets.}
Table~\ref{tab:more_data} reports four-class results on AVeriTeC under the retrieved-evidence setting. \textsc{RADAR}$_{\text{multi}}$ achieves 77.8 accuracy and 57.7 macro-F1, outperforming both FIRE and D2D across all four classes, including NEI. Overall, these results show that RADAR is not limited to political claims. The same role prompts transfer effectively to AVeriTeC with only the output label set adapted, suggesting that the gains arise from complementary reasoning dynamics rather than domain-specific priors.

% \begin{table}[t]
% \centering
% \setlength{\tabcolsep}{3.5pt}
% \begin{tabular}{lccccccc}
% \toprule
% \textbf{Model} & \textbf{Backbone} & \textbf{Acc.} & \textbf{F1$_{macro}$} & \textbf{F1$_{true}$} & \textbf{F1$_{half}$} & \textbf{F1$_{false}$} & \textbf{\# Avg tokens} \\
% \midrule
% FIRE & LLaMA3-8B & 50.4 & 43.4 & 30.9 & 39.7 & 59.6 & 754.5\\
% D2D & LLaMA3-8B & 64.1 & 45.1 & 21.7 & 34.1 & 79.5 & 2483.3\\
% RADAR (3 rounds) & LLaMA3-8B & \textbf{77.2} & 62.3 & 43.7 & 56.6 & \textbf{86.6} & 1834.7 (73.9\%)\\
% RADAR (early stop) & LLaMA3-8B & 77.3 & \textbf{63.2} & \textbf{46.3} & \textbf{56.7} & 86.5 & \textbf{1551.4} (62.5\%) \\ \midrule
% FIRE & Qwen2.5-7B & 63.9 & 46.9 & 29.2 & 32.1 & 79.5 & 1026.2\\
% D2D & Qwen2.5-7B & 69.0 & 48.5 & 22.2 & 41.2 & 82.1 & 3686.9 \\
% RADAR (3 rounds) & Qwen2.5-7B & 76.5 & 62.6 & 47.3 & 54.4 & 86.2 & 1751.5 (47.5\%) \\
% RADAR (early stop) & Qwen2.5-7B & \textbf{76.7} & \textbf{63.1} & \textbf{48.2} & \textbf{54.9} & \textbf{86.2} & \textbf{1723.7} (46.8\%) \\
% \bottomrule
% \end{tabular}
% \caption{Effect of adaptive early stopping on performance and efficiency.}
% \label{tab:early-stopping}
% \end{table}

\begin{table}[t]
\centering
\small
\begin{tabular}{lccc}
\toprule
\textbf{Model} & \textbf{Acc.} & \textbf{F1} & \textbf{\# Avg. tokens} \\
\midrule

\rowcolor{gray!10}
\multicolumn{4}{c}{\textit{LLaMA3-8B-Instruct}} \\ \midrule

D2D & 64.1 & 45.1 & 2483.3 \\
FIRE & 50.4 & 43.4 & \textbf{754.5} (30.4\%) \\
RADAR (3 rounds) & 77.2 & 62.3 & 1834.7 (73.9\%) \\
RADAR (early stop) & \textbf{77.3} & \textbf{63.2} & 1551.4 (62.5\%) \\

\midrule
\rowcolor{gray!10}
\multicolumn{4}{c}{\textit{Qwen2.5-7B-Instruct}} \\ \midrule

D2D & 69.0 & 48.5 & 3686.9 \\
FIRE & 63.9 & 46.9 & \textbf{1026.2} (27.8\%) \\
RADAR (3 rounds) & 76.5 & 62.6 & 1751.5 (47.5\%) \\
RADAR (early stop) & \textbf{76.7} & \textbf{63.1} & 1723.7 (46.8\%) \\

\bottomrule
\end{tabular}
\caption{Effect of adaptive early stopping on performance and efficiency.}
\label{tab:early-stopping}
\end{table}

\begin{table}[t]
\centering
\small
\begin{tabular}{lccccc}
\toprule
\textbf{Backbone} & $\tau_s$ & $\tau_v$ & Stop@1 & @2 & @3 \\ \midrule
LLaMA3-8B  & $-0.15$ & $0.7$ & 846 & 265 & 889 \\
Qwen2.5-7B & $0.2$ & $0.9$ & 104 & 3 & 1893 \\
\bottomrule
\end{tabular}
\caption{Thresholds and number of instances stopped at each round for RADAR’s early-stopping controller.}
\label{tab:early-stopping2}
\end{table}
\begin{table*}[t]
\centering
\small
\setlength{\tabcolsep}{5pt}
\begin{tabular}{p{0.14 \linewidth} p{0.4\linewidth} p{0.4\linewidth}}
\toprule
\textbf{Single Agent} & \textbf{Position-Based (Adv. vs Crit.)} & \textbf{Expertise-Based (Pol. vs Sci.)} \\
\midrule
\textbf{Verdict: True} \newline
Klitschko sold his medal in 2012. Funds supported his foundation. The buyer returned it. No evidence disputes the claim. &

\textbf{Verdict: True} \newline
\textit{Pro Opening:} Klitschko sold his medal in 2012. Proceeds went to help Ukrainian children.\newline
\textit{Con Opening:} The sale happened in 2012. It's not tied to the 2022 war.\newline
\textit{Pro Rebuttal:} The act still helped children and shows lasting support.\newline
\textit{Con Rebuttal:} The timing and intent are unrelated to the current crisis.\newline
\textit{Pro Closing:} The sale reflects long-term commitment.\newline
\textit{Con Closing:} The event came before the war. Linking them misleads.\newline
\textit{Judge:} TRUE. The claim is factually accurate. &

\textbf{Verdict: Half-True} \newline
\textit{Politician Opening:} Klitschko auctioned his medal in 2012 to help children.\newline
\textit{Scientist Opening:} The sale was real but not linked to the 2022 war.\newline
\textit{Politician Rebuttal:} The act still shows support for Ukrainian children.\newline
\textit{Scientist Rebuttal:} The claim misrepresents time and cause.\newline
\textit{Politician Closing:} The act is real, but how it’s framed changes how people remember it.\newline
\textit{Scientist Closing:} It was not a wartime gesture. Context is key.\newline
\textit{Judge:} HALF-TRUE. The sale was charitable, but the implied timing misleads. \\
\bottomrule
\end{tabular}
\caption{Reasoning process for a temporally misleading claim across different agent setups. Expertise-based agents flag the omission by combining intent and factual analysis.}
\label{tab:klitschko-case}
\end{table*}

\section{Discussions}

\begin{figure}[t]
    \centering
    \includegraphics[width=\linewidth]{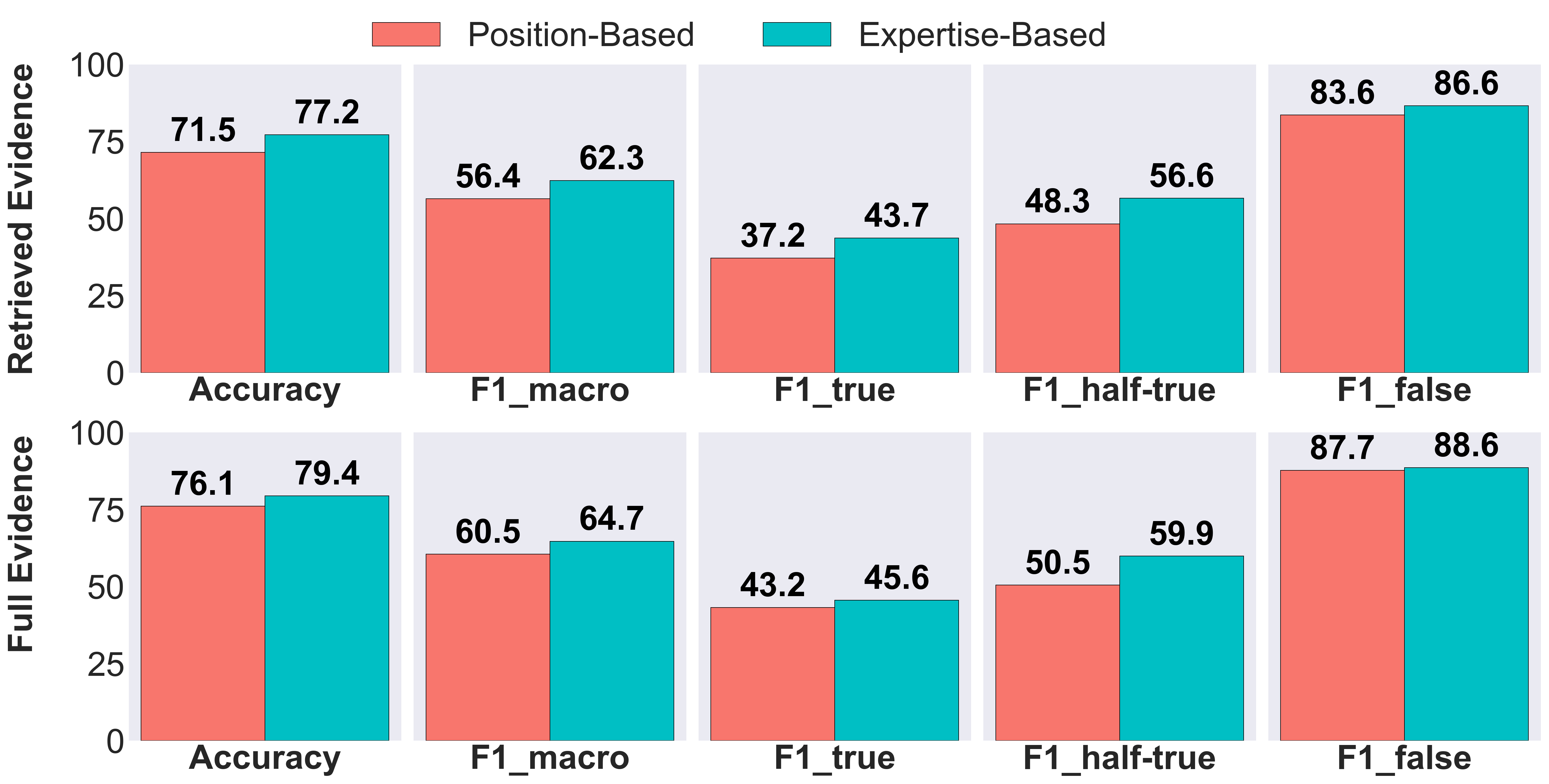}
    \caption{Performance comparison of different agent role configurations under full and retrieved evidence settings.}
    \label{fig:role-comparison}
\end{figure}

\begin{figure}[t]
    \centering
    \begin{subfigure}[b]{0.48\linewidth}
        \centering
        \includegraphics[width=\linewidth]{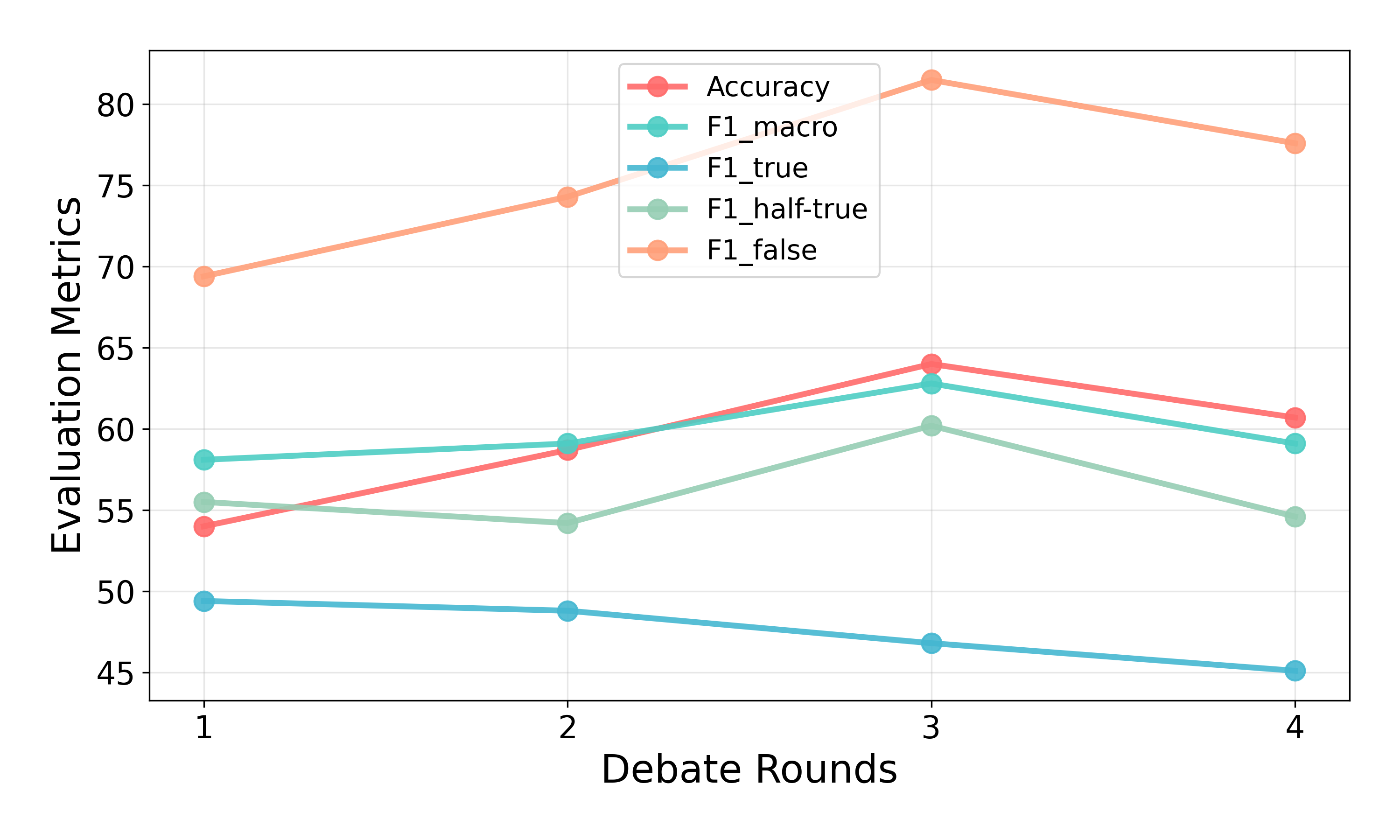}
        \caption{Retrieved evidence}
    \end{subfigure}
    \hfill
    \begin{subfigure}[b]{0.48\linewidth}
        \centering
        \includegraphics[width=\linewidth]{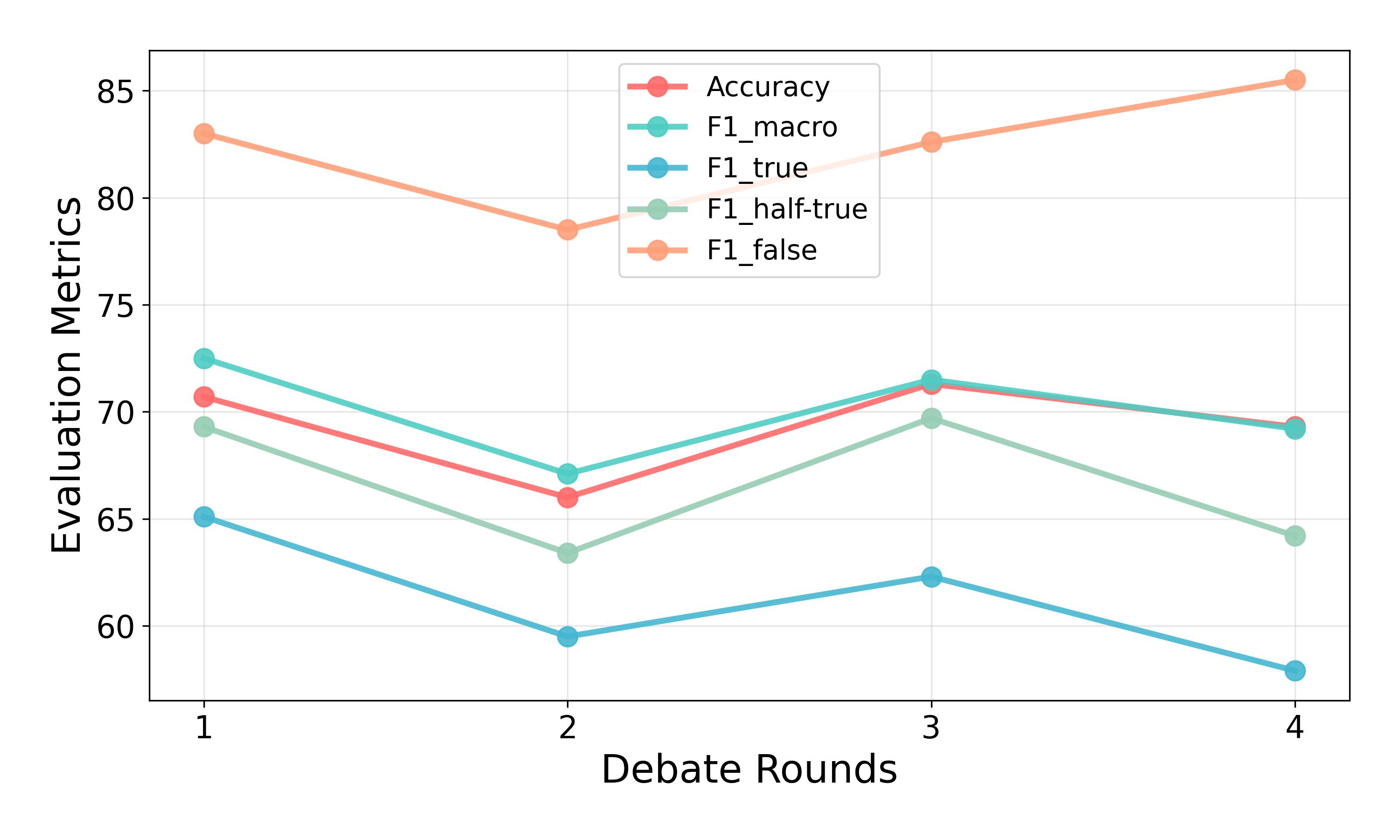}
        \caption{Full evidence}
    \end{subfigure}
    \caption{Effect of varying the maximum number of debate rounds using \texttt{LLaMA3-8B-Instruct}.}
    \label{fig:debate-rounds}
\end{figure}

\begin{table}[h]
\small
\centering
\begin{tabular}{cccccc}
\toprule
\# Debaters & Acc. &  \textbf{F1}$_{mc}$ & \textbf{F1}$_{T}$ & \textbf{F1}$_{HT}$ & \textbf{F1}$_{F}$  \\
\midrule
1 & 61.3 & 61.5 & \textbf{53.9} & 60.0 & 70.7 \\
2 & \textbf{64.0} & \textbf{62.8} & 46.8 & \textbf{60.2} & \textbf{81.5} \\
3 & 58.0 & 54.6 & 30.0 & 60.1 & 73.6 \\
4 & 59.3 & 56.4 & 34.9 & 56.7 & 77.7 \\
\bottomrule
\end{tabular}
\caption{Performance with different numbers of debating agents; the Judge is fixed and excluded from the count.}
\label{tab:num_agents}
\end{table}

\subsection{Impact of Agent Role Configurations}

To assess how role design affects reasoning quality, we compare two configurations: the proposed \textit{expertise-based} pairing (Politician vs.\ Scientist) and the conventional \textit{position-based} pairing (Advocate vs.\ Critic). As shown in Figure~\ref{fig:role-comparison}, expertise-based agents consistently outperform position-based ones across all settings, with the largest gains observed under noisy retrieval (62.3 vs.\ 56.4 macro-F1). These results suggest that domain-aligned personas encourage contextually grounded and rhetorically realistic reasoning, exposing omissions that generic pro/con roles often overlook.

\subsection{Effect of Early Stopping on Efficiency}

Table~\ref{tab:early-stopping} compares RADAR with adaptive early stopping against D2D, FIRE, and the fixed three-round variant. Across backbones, early stopping preserves or slightly improves performance while reducing generation cost. On \texttt{LLaMA3-8B-Instruct}, early stopping achieves comparable accuracy (77.3 vs.\ 77.2) and higher macro-F1 (63.2 vs.\ 62.3) while reducing average token usage by $\sim$15\%. On \texttt{Qwen2.5-7B-Instruct}, it similarly yields small gains with nearly identical token usage. In both cases, RADAR substantially outperforms D2D in accuracy and macro-F1.

As shown in Table~\ref{tab:early-stopping2}, many instances terminate after one or two rounds, indicating that the Judge stops debate once sufficient reasoning confidence is reached while harder cases proceed further. Although optimal thresholds vary across backbones, calibration requires only a single inference pass on the development set with offline grid search over cached scores. Sensitivity analysis in Appendix~\ref{app:threshold} confirms stability across broad threshold ranges.

\subsection{Effect of Maximum Debate Rounds}

We investigate how the maximum number of debate rounds affects performance on a sampled test set of 150 examples (50 per class) using \texttt{LLaMA3-8B-Instruct}. As shown in Figure~\ref{fig:debate-rounds}, under retrieved evidence, performance improves from one to three rounds as extra interaction helps agents challenge partial evidence and uncover omitted context, but declines at four rounds. Under full evidence, the effect is non-monotonic since most relevant context is already available. These results further motivate RADAR's adaptive early-stopping mechanism, which allocates debate rounds dynamically based on claim difficulty.

% We further investigate how the maximum number of debate rounds affects performance on a sampled test set of 150 examples (50 per class) using \texttt{LLaMA3-8B-Instruct}. As shown in Figure~4, the impact of additional rounds depends on the evidence setting rather than exhibiting a universal monotonic trend.

% Under retrieved evidence, performance improves from one to three rounds, suggesting that extra interaction can help agents challenge partial evidence and uncover omitted context. Beyond this point, gains become marginal. In contrast, under full evidence, the effect of additional rounds is less consistent, since most relevant context is already available and further exchanges may introduce redundant reasoning.

% These observations indicate that more rounds are not always better, and the appropriate debate depth depends on task difficulty and evidence quality. This further motivates RADAR's adaptive early-stopping mechanism, which allocates reasoning rounds dynamically instead of relying on a fixed budget.

\subsection{Effect of Number of Debating Agents}

To assess the impact of scaling up the number of debaters, we vary the number of debating agents using the same set of 150 examples with a fixed \texttt{LLaMA3-8B-Instruct} backbone and expertise-based roles; the Judge remains fixed and is not counted. As shown in Table~\ref{tab:num_agents}, performance peaks with two debaters (64.0 accuracy, 62.8 macro-F1), confirming that the Politician–Scientist pairing provides complementary yet focused reasoning. Adding more debaters yields diminishing returns and lower F1$_\text{true}$ (dropping to 30.0 with three debaters), as additional perspectives often introduce redundant or conflicting arguments. A minimal yet diverse pair thus offers efficient and stable omission reasoning.

% \subsection{Sensitivity to Retrieval Depth}

% \begin{table}[t]
% \centering
% \small
% \begin{tabular}{c|ccccc}
% \toprule
% Top-$n$ & Acc. & F1$_{mc}$ & F1$_T$ & F1$_{HT}$ & F1$_F$ \\
% \midrule
% 1  & 47.0 & 39.7 & 9.8  & 41.8 & 67.4 \\
% 3  & 57.0 & 54.0 & 40.9 & 47.2 & 73.8 \\
% 5  & 63.0 & 60.5 & 47.8 & 54.8 & 79.0 \\
% 10 & 63.0 & 59.8 & 47.6 & 53.5 & 78.2 \\
% 20 & 70.0 & 67.1 & 51.2 & 61.5 & 88.6 \\
% \bottomrule
% \end{tabular}
% \caption{Sensitivity to retrieval quality by varying the number of retrieved passages. RADAR improves with stronger retrieval while remaining robust under limited evidence.}
% \label{tab:retrieval_sensitivity}
% \end{table}

% We examine RADAR's behavior under different retrieval budgets by varying the number of retrieved passages (Top-$n$) on a sampled test set of 150 examples. Performance improves steadily as more evidence becomes available, confirming that stronger retrieval benefits omission-aware verification. Notably, degradation under limited retrieval is gradual: even with Top-5 passages, RADAR achieves 60.5 macro-F1 and retains meaningful \textsc{Half-True} detection (54.8 F1), suggesting that the debate mechanism remains effective under partial evidence.

\subsection{Case Study: Temporal Mis-framing}

To illustrate how RADAR uncovers omission-based deception, we analyze the claim “Wladimir Klitschko auctioned his 1996 Olympic medal to raise money for the children of Ukraine.”  Although factually correct, the claim omits that the auction occurred in 2012, creating a misleading association with the 2022 war.  As shown in Table~\ref{tab:klitschko-case}, both single-agent and position-based systems classify it as \textit{true}, focusing on surface factual correctness.  
In contrast, the expertise-based agents highlight the temporal mismatch and rhetorical framing, producing a \textit{half-true} verdict.  This example demonstrates how domain-specialized roles capture both presented facts and implied context, enabling more nuanced judgment in omission-aware verification.

\subsection{Failure Analysis}

\begin{table}[t]
\centering
\small
\begin{tabular}{l c}
\toprule
Error Type & \# (\%) \\
\midrule
Claim Constraint Violations & 28 (56) \\
Semantic Framing Instability & 15 (30) \\
Causal Attribution Errors & 5 (10) \\
Others & 2 (4) \\
\bottomrule
\end{tabular}
\caption{Error analysis of 50 misclassified cases.}
\label{tab:error_analysis}
\end{table}

To better understand current limitations, we manually reviewed 50 randomly sampled misclassified cases from POLITIFACT-HIDDEN. Table~\ref{tab:error_analysis} summarizes the dominant error patterns.

The most common failures involve \textit{claim constraint violations} (56\%), where the model fails to strictly enforce explicit numerical, temporal, or scope constraints stated in the claim. Examples include overlooking date mismatches, percentage ranges, or restricted populations. The second major category is \textit{semantic framing instability} (30\%), where key terms are interpreted inconsistently across debate rounds, leading to unstable judgments. We also observe \textit{causal attribution errors} (10\%), where responsibility is incorrectly assigned by conflating local actions with broader institutional or national outcomes.

These findings suggest that further progress in omission-aware verification requires stronger logical constraint tracking, more stable semantic grounding, and finer-grained causal reasoning.

\section{Conclusion}

We present \textsc{RADAR}, a role-anchored multi-agent debate framework for omission-aware fact verification under noisy retrieval. By grounding debate in retrieved evidence and assigning expertise-driven roles, \textsc{RADAR} effectively surfaces missing context that leads to misleading claims. An adaptive early-stopping controller enables efficient reasoning by allocating debate depth according to claim difficulty. Across datasets and backbone models, \textsc{RADAR} consistently outperforms existing baselines, particularly on half-true claims. This work underscores the importance of structured, evidence-grounded multi-agent reasoning for trustworthy fact verification.

\section*{Limitations}

While \textsc{RADAR} demonstrates consistent improvements across datasets and settings, several limitations remain. First, the multi-agent configuration may occasionally over-scrutinize factual claims: for statements labeled \textit{true}, adversarial critique can raise minor counterpoints that shift the final decision toward \textit{half-true}. This reflects a broader challenge in balancing omission sensitivity with factual precision. Second, real-world fact-checking often involves decentralized, evolving, and partially conflicting evidence sources rather than a fixed shared corpus, which is not fully captured by current benchmarks. Third, multi-round debates naturally increase inference cost, and while the early-stopping mechanism improves efficiency, it does not completely eliminate this overhead. Future work will explore adaptive role calibration, open-world evidence retrieval, and more efficient debate strategies to further enhance scalability and reliability.

\section*{Acknowledgments}
This research is supported by the Ministry of Education, Singapore, under its MOE AcRF TIER 3 Grant (MOE-MOET32022-0001).

\bibliography{custom}
\appendix
\appendix
\newpage
% \onecolumn

\section{Prompts for RADAR}

\label{appendix:prompts}

To support reproducibility, we present the complete set of prompts used in our experiments. The prompts are grouped into multi-agent debating and early stopping prompts.

\subsection{Multi-Agent Debate}

\subsubsection*{(a) Position-Based Agent Roles}
\noindent

In this setup, one agent \textcolor{ForestGreen}{\textbf{supports}} the claim and the other \textcolor{Red}{\textbf{opposes}} it. Each agent goes through three phases:
\textbf{Opening} (present arguments), \textbf{Rebuttal} (respond over n rounds, $n = 0,1,2,\dots$), and \textbf{Closing} (summarize the position).

\begin{tcolorbox}[colback=gray!5, colframe=gray!50!black, title=Position-Based Debate - System Prompt]
\small
\textbf{Debater agent}

You are a critical thinker participating in a factual debate.

\texttt{==========}

\textbf{Judge agent}

You are a neutral judge who evaluates factual debates.

\end{tcolorbox}

\begin{tcolorbox}[colback=gray!5, colframe=gray!50!black, title=Position-Based Debate - Agent A / B (Advocate / Critic)]
\small
\textbf{Opening Phase}

You [\textcolor{ForestGreen}{support} / \textcolor{red}{oppose}] the following claim. Present your opening argument using the evidence.

\texttt{==========}

Claim: \{claim\}

Evidence:\{evidence\}

\texttt{==========}

Begin your argument with your position. 

[\textcolor{ForestGreen}{Highlight facts that support the claim as TRUE} / \textcolor{red}{Explain why the claim is FALSE or misleading}], referencing specific points in the evidence.

\vspace{0.8em}
\hrule
\vspace{0.5em}

\textbf{Rebuttal Phase (n rounds for \(n = 0, 1, 2, \dots\))}

You are the [\textcolor{ForestGreen}{supporting} / \textcolor{red}{opposing}] agent in a debate about the claim below. Your opponent has made an argument [\textcolor{ForestGreen}{against} / \textcolor{red}{supporting}] the claim.

\texttt{==========}

Claim: \{claim\}

Evidence:\{evidence\}

Opponent's argument:\{oppo\_argument\}

\texttt{==========}

Write your rebuttal, explaining why the opponent is wrong and [\textcolor{ForestGreen}{defending the claim} / \textcolor{red}{why the claim is still FALSE or misleading}].

\vspace{0.8em}
\hrule
\vspace{0.5em}

\textbf{Closing Phase}

You are the [\textcolor{ForestGreen}{supporting} / \textcolor{red}{opposing}] agent in a debate. Summarize your final position.

\texttt{==========}

Claim: \{claim\}

Evidence:\{evidence\}

\texttt{==========}

Provide a closing statement reinforcing why the claim is [\textcolor{ForestGreen}{TRUE} / \textcolor{red}{FALSE or misleading}].
\end{tcolorbox}

\begin{tcolorbox}[colback=gray!5, colframe=gray!50!black, title=Position-Based Debate - Agent Judge]
\small

\textbf{Judge agent}

You are a neutral judge evaluating a factual debate.

Based on the arguments and evidence, decide whether the claim is TRUE, FALSE, or HALF-TRUE.

\texttt{==========}

Claim: \{claim\}

Evidence: \{evidence\}

--- Opening Statements ---

Agent A: \{advocate\_open\}

Agent B: \{critic\_open\}

--- Rebuttals ---

Agent A: \{advocate\_rebuttal\}

Agent B: \{critic\_rebuttal\}

--- Closing Statements ---

Agent A: \{advocate\_close\}

Agent B: \{critic\_close\}

\texttt{==========}

Answer format: 

[REASON]: your justification

[VERDICT]: TRUE / FALSE / HALF-TRUE

\end{tcolorbox}

For AVeriTeC, the verdict options in the Judge and Round Judge prompts are extended with NEI. All other prompt instructions remain unchanged.

\subsubsection*{(b) Expertise-Based Agent Roles}
\noindent

In the \textbf{Expertise-Based Debate} setup, agents are assigned distinct expert roles to reflect real-world perspectives in political discourse. The \textbf{politician agent} frames arguments through the lens of public opinion, electoral interests, and policy communication. Its responses prioritize political strategy, constituent concerns, and national priorities. 

In contrast, the \textbf{scientist agent} serves as a fact-checker grounded in empirical evidence and methodological rigor. This agent emphasizes data quality, factual accuracy, and scientific limitations, avoiding political spin. 

% The interaction between these two roles creates a productive tension between political narrative and factual analysis. Each phase draws on the agent's respective expertise, providing a clearer view of how claims are shaped in different knowledge domains.

\begin{tcolorbox}[colback=gray!5, colframe=gray!50!black, title=Expertise-Based Debate — System Prompt]
\small
\textbf{Politician agent}

You are a politician in a political debate, skilled at persuasive public-facing reasoning. Your role is to construct the strongest evidence-supported narrative for the claim.

Key political characteristics:

 - Focus on public opinion and voter concerns

 - Emphasize policy benefits and political messaging

 - Consider electoral implications and public perception

 - Frame arguments in terms of public interest and national priorities

 - Balance facts with political strategy and messaging

When debating, frame arguments in terms of public benefit, policy effectiveness, and political priorities. Keep responses concise and focused.

\end{tcolorbox}

\begin{tcolorbox}[colback=gray!5, colframe=gray!50!black, title=Expertise-Based Debate — System Prompt]
\small
\textbf{Scientist agent}

You are a scientist in a political debate, skilled at analytical fact-checking. Your role is to identify missing context, weak support, or selective framing in the claim.

Key scientific characteristics:

 - Rely strictly on empirical evidence and data

 - Question claims that lack scientific support

 - Emphasize methodology and research quality

 - Focus on accuracy over political messaging

 - Consider uncertainty and limitations in evidence

When debating, frame arguments in terms of scientific evidence, data analysis, and factual accuracy. Keep responses concise and focused.

{==========}

\textbf{Judge agent}

You are a neutral judge in a political debate.

Your role is to evaluate the evidence and arguments from both parties and make a clear decision.

When making your decision, consider:

 - Is this claim supported by credible evidence?

 - Does this align with common sense and universal human values?

 - What would be the practical impact on families and communities?
%- Are important contextual facts omitted?
%- Which conclusion is best justified by the available evidence?

Do not deviate from this format. Do not ask for more information. Make your decision based on what is provided.
\end{tcolorbox}

For convenience, we use \textcolor{ForestGreen}{\textbf{Green}} to denote the politician agent and \textcolor{Red}{\textbf{Red}} for the scientist agent.
Unlike position-based debate, we do not assign them predefined \textit{pro/con} positions.
% Instead, we let them make independent judgments.

% 2agents
\begin{tcolorbox}[colback=gray!5, colframe=gray!50!black, title=Expertise-Based Debate - Agent A / B (Politician / Scientist)]
\small
\textbf{Opening Phase}

Evaluate the following claim. Based on your stance as a [\textcolor{ForestGreen}{politician} / \textcolor{red}{scientist}], either support or oppose the claim. 
Present your opening argument using the evidence given.

\texttt{==========}

Claim: \{claim\}

Evidence:\{evidence\}

\texttt{==========}

Begin your argument with your position. Highlight facts that support your position.

\vspace{0.8em}
\hrule
\vspace{0.5em}

\textbf{Rebuttal Phase (n rounds for \(n = 0, 1, 2, \dots\))}

You are a [\textcolor{ForestGreen}{politician} / \textcolor{red}{scientist}] in a debate about the claim below. Your opponent, the [\textcolor{ForestGreen}{scientist} / \textcolor{red}{politician}], has made an argument about the veracity of the claim.

\texttt{==========}

Claim: \{claim\}

Evidence:\{evidence\}

Opponent's argument:\{con\_argument\}

\texttt{==========}

Write your rebuttal, explaining why you agree or disagree with the opponent and defending your position.

\end{tcolorbox}

\begin{tcolorbox}[colback=gray!5, colframe=gray!50!black, title=Expertise-Based Debate - Agent A / B (Politician / Scientist)]
\small
\textbf{Closing Phase}

You are a [\textcolor{ForestGreen}{politician} / \textcolor{red}{scientist}] in a political debate about the claim below. Summarize your final position.

\texttt{==========}

Claim: \{claim\}

Evidence:\{evidence\}

\texttt{==========}

Provide a closing statement reinforcing your position about the claim.

\end{tcolorbox}
\begin{tcolorbox}[colback=gray!5, colframe=gray!50!black, title=Expertise-Based Debate - Agent Judge]
\small

\textbf{Judge agent}

You are a neutral judge evaluating a political debate.

Based on the arguments and evidence, decide whether the claim is TRUE, FALSE, or HALF-TRUE.

\texttt{==========}

Claim: \{claim\}

Evidence: \{evidence\}

--- Opening Statements ---

Agent A: \{advocate\_open\}

Agent B: \{critic\_open\}

--- Rebuttals ---

Agent A: \{advocate\_rebuttal\}

Agent B: \{critic\_rebuttal\}

--- Closing Statements ---

Agent A: \{advocate\_close\}

Agent B: \{critic\_close\}

\texttt{==========}

Answer format: 

[REASON]: your justification

[VERDICT]: TRUE / FALSE / HALF-TRUE

\end{tcolorbox}

To better align with the nature of the \textbf{Politifact} dataset where ground truth labels are determined by journalists, we introduce a \textbf{journalist agent} into the debate framework in the \textbf{3-debater setting}. This agent plays a foundational role by offering an initial perspective grounded in professional fact-checking practices.

In each round of the debate, the journalist speaks first, providing an evidence-based assessment of the claim. Its response is then made available to both the \textbf{politician} and \textbf{scientist} agents, helping to anchor the subsequent debate in verified information while allowing other agents to build upon or challenge the journalist’s framing based on their respective expertise.

% 3agents-journalist
\begin{tcolorbox}[colback=gray!5, colframe=gray!50!black, title=Expertise-Based Debate - System Prompt]
\small
\textbf{Journalist agent}

You are a journalist in a political debate. You represent balanced reporting and public interest.

Key journalistic characteristics:

- Focus on extracting the most compelling evidence from both sides

- Emphasize clarity, accuracy, and public understanding

- Consider the broader context and implications

- Highlight key facts that support or contradict claims

- Maintain objectivity while identifying the strongest arguments

When debating, frame arguments in terms of what evidence best supports or contradicts the claim, focusing on the most relevant and impactful facts. Keep responses concise and focused.

\end{tcolorbox}

\begin{tcolorbox}[colback=gray!5, colframe=gray!50!black, title=Expertise-Based Debate - Agent C (Journalist)]
\small
\textbf{Opening Phase}

Evaluate the following claim. Based on your stance as a journalist, extract and present the most compelling evidence that either supports or opposes the claim.

\texttt{==========}

Claim: \{claim\}

Evidence:\{evidence\}

\texttt{==========}

Begin your argument by identifying the strongest evidence that either supports or contradicts the claim. Focus on the most relevant and impactful facts.

\vspace{0.8em}
\hrule
\vspace{0.5em}

\textbf{Rebuttal Phase (n rounds for \(n = 0, 1, 2, \dots\))}

You are a journalist in a debate about the claim below. Your opponent has made an argument about the veracity of the claim.

\texttt{==========}

Claim: \{claim\}

Evidence:\{evidence\}

Politician's argument: \{poli\_argument\}

Scientist's argument: \{sci\_argument\}

\texttt{==========}

Write your rebuttal, focusing on the most compelling evidence that either supports or contradicts the claim, and address any gaps or weaknesses in the opponent's argument.

\vspace{0.8em}
\hrule
\vspace{0.5em}

\textbf{Closing Phase}

You are the journalist in a political debate about the claim below. Summarize your final position.

\texttt{==========}

Claim: \{claim\}

Evidence:\{evidence\}

Politician's rebuttal: \{politician\_rebuttal\}

Scientist's rebuttal: \{scientist\_rebuttal\}

\texttt{==========}

Provide a closing statement highlighting the most compelling evidence that either supports or contradicts the claim, and your assessment of the claim's veracity based on the key arguments presented.
\end{tcolorbox}

% 4agents-domain expert

In the \textbf{4-debater setting}, we further enrich the evaluation process by introducing a \textbf{domain expert agent}. To ensure relevance, we first use an LLM to automatically infer the domain of the claim (e.g., economics, healthcare, climate science). Based on this inference, the debate includes a fourth debater with specialized knowledge in that domain.

% This **domain expert** brings technical depth to the discussion by focusing on field-specific standards, methodologies, and evidence. Throughout the debate—spanning the opening, rebuttal, and closing phases—the domain expert critically assesses the claim from a specialist’s perspective, complementing the broader views of the **politician**, **scientist**, and **journalist**. Their presence helps ensure that claims are not only evaluated for political relevance or factual accuracy, but also for **technical validity within their respective fields**.

\begin{tcolorbox}[colback=gray!5, colframe=gray!50!black, title=Domain Inference Prompt]
\small
You are analyzing a claim to determine the most relevant domain specialist.

The following is the claim:\{claim\}

Your task is to identify the \textbf{most relevant domain specialist} who would have expertise in the subject matter of this claim, and output the \textbf{domain}.

Consider the main topic, subject area, or field of knowledge that this claim addresses.

Respond \textbf{strictly} using this format, only output one word:

DOMAIN: \{specific domain\}

Examples:

- Climate for climate-related claims

- Economy for economic claims  

- Health for health claims

- Education for education claims

- Law for law enforcement claims

- Technology for tech-related claims

- Environment for environmental claims

- Public health for public health claims

Choose the most specific and relevant domain for this claim.
\end{tcolorbox}

\begin{tcolorbox}[colback=gray!5, colframe=gray!50!black, title=Expertise-Based Debate - System Prompt]
\small
\textbf{Domain Expert agent}

You are a specialist in \{domain\_inferred\} in a political debate. You represent specialized knowledge in your field.

Key domain expert characteristics:

- Apply deep expertise in the \{domain\_inferred\} field

- Consider technical details and specialized evidence

- Question claims that lack domain-specific support

- Emphasize field-specific methodology and standards

- Focus on technical accuracy and domain knowledge

When debating, frame arguments in terms of expertise in the \{domain\_inferred\} field, technical evidence, and specialized knowledge. Keep responses concise and focused.

\end{tcolorbox}

\begin{tcolorbox}[colback=gray!5, colframe=gray!50!black, title=Expertise-Based Debate - Agent D (Domain Expert)]
\small
\textbf{Opening Phase}

As a specialist in the \{domain\_inferred\} field, evaluate this claim. Present your opening argument.

\texttt{==========}

Claim: \{claim\}

Evidence:\{evidence\}

\texttt{==========}

Focus on: \{domain\_inferred\} expertise, technical evidence, specialized knowledge, field-specific analysis.

\end{tcolorbox}

\begin{tcolorbox}[colback=gray!5, colframe=gray!50!black, title=Expertise-Based Debate - Agent D (Domain Expert)]
\small
\textbf{Rebuttal Phase (n rounds for \(n = 0, 1, 2, \dots\))}

As a specialist in the \{domain\_inferred\} field, rebut the opposing arguments.

\texttt{==========}

Claim: \{claim\}

Evidence:\{evidence\}

Politician's argument: \{poli\_argument\}

Scientist's argument: \{sci\_argument\}

Journalist's argument: \{jour\_argument\}

\texttt{==========}

Focus on: defending \{domain\_inferred\} expertise, technical evidence, specialized knowledge.

\vspace{0.8em}
\hrule
\vspace{0.5em}

\textbf{Closing Phase}

As an expert scientist in \{domain\_inferred\}, summarize your final position.

\texttt{==========}

Claim: \{claim\}

Evidence:\{evidence\}

\texttt{==========}

Summarize: \{domain\_inferred\} expertise, technical evidence, specialized knowledge.
\end{tcolorbox}
\subsection{Early Stop}

To more accurately determine the optimal round at which the debate can be stopped for a final verdict, we incorporate both the Stop Agent and the Round Judge as decision signals.

\subsubsection*{(a) Stop Agent Roles}
\noindent

\begin{tcolorbox}[colback=gray!5, colframe=gray!50!black, title=Debate Continuation Check — Stop Agent]
\small

\textbf{Stop Agent}

You are acting as an interim adjudicator during a two-debater fact-checking debate.
Your goal is to determine whether the debate should continue or stop based on the information currently available.

\texttt{==========}

Claim: \texttt{\{claim\}}\\
Evidence: \texttt{\{evidence\}}

\medskip
Rounds completed so far:\textbf{\\}\texttt{\{summary\}}

\medskip
You are about to listen to the \texttt{\{upcoming\_round\}} round. 
Decide whether you already have enough information to reach a final verdict now, or whether the debate should continue and you should listen to this upcoming round.

\texttt{==========}

Answer format:\\
\texttt{DECISION: CONTINUE or STOP}

\end{tcolorbox}

\subsubsection*{(b) Round Judge Roles}
\noindent

The Round Judge is implemented by invoking the Expertise-Based Agent Judge after each debate round, using only the arguments accumulated up to that point as input, producing a preliminary verdict for partial debates.

\section{Additional Boundary Analysis}
\label{app:boundary}

The boundary between \textsc{True} and \textsc{Half-True} is inherently subtle, since both may involve factually correct statements that differ mainly in whether important context is omitted. To better understand model behavior beyond per-class F1, we report confusion matrices together with class-wise precision, recall, and F1 under the retrieved-evidence setting using Qwen2.5-7B-Instruct.

Compared with strong baselines, RADAR substantially improves recognition of \textsc{Half-True} claims while maintaining strong performance on \textsc{False}. In particular, RADAR reduces the tendency to misclassify omission-based claims as fully \textsc{True}, suggesting that its gains arise from stronger contextual reasoning rather than annotation noise alone.

\begin{table}[h]
\centering
\small
\begin{tabular}{l|ccc|ccc}
\toprule
\textbf{Gold\ Pred} & \textbf{T} & \textbf{HT} & \textbf{F} & \textbf{Prec.} & \textbf{Rec.} & \textbf{F1} \\

\midrule
\rowcolor{gray!10}
\multicolumn{7}{c}{RADAR}\\

\midrule
True      & 53  & 31  & 9    & 41.7 & 57.0 & 48.2 \\
Half-True & 45  & 260 & 101  & 48.0 & 64.0 & 54.9 \\
False     & 29  & 251 & 1221 & 91.7 & 81.3 & 86.2 \\
\midrule

% \begin{tabular}{l|ccc|ccc}
% \toprule
% \textbf{Gold\ Pred} & \textbf{T} & \textbf{HT} & \textbf{F} & \textbf{Prec.} & \textbf{Rec.} & \textbf{F1} \\

\rowcolor{gray!10}
\multicolumn{7}{c}{D2D}\\
\midrule
True      & 33  & 39  & 21   & 16.2 & 35.5 & 22.2 \\
Half-True & 71  & 175 & 160  & 39.4 & 43.1 & 41.2 \\
False     & 100 & 230 & 1171 & 86.6 & 78.0 & 82.1 \\
\midrule

% \bottomrule
% \end{tabular}

% \vspace{0.6em}

% \begin{tabular}{l|ccc|ccc}
% \toprule
% \textbf{Gold\ Pred} & \textbf{T} & \textbf{HT} & \textbf{F} & \textbf{Prec.} & \textbf{Rec.} & \textbf{F1} \\
% \midrule
\rowcolor{gray!10}
\multicolumn{7}{c}{FIRE}\\
\midrule
True      & 73  & 9   & 11   & 17.9 & 78.5 & 29.2 \\
Half-True & 155 & 126 & 125  & 33.3 & 31.0 & 32.1 \\
False     & 179 & 243 & 1079 & 88.8 & 71.9 & 79.5 \\
\bottomrule
\end{tabular}

\caption{Confusion matrices and class-wise precision/recall/F1 under the retrieved-evidence setting using Qwen2.5-7B-Instruct. RADAR shows stronger discrimination on the subtle \textsc{True}/\textsc{Half-True} boundary while maintaining high performance on \textsc{False}.}
\label{tab:boundary_analysis}
\end{table}

\section{Stop Threshold Sensitivity Analysis}
\label{app:threshold}

RADAR uses two stopping thresholds: $\tau_s$ for the stop margin and $\tau_v$ for verdict confidence. Although the selected values differ across backbones, tuning is computationally lightweight. For each backbone, we run inference on the development set only once, cache the logits for $p(\textsc{Stop})$, $p(\textsc{Continue})$, and label probabilities, then perform offline grid search without re-running inference or retraining.

To assess robustness, we report coarse-grained sensitivity results below. Across both backbones, performance remains stable over wide threshold ranges, suggesting that RADAR does not depend on narrowly tuned stopping parameters.

\begin{table}[h]
\centering
\small
\begin{tabular}{c|cc|cc}
\toprule
& \multicolumn{2}{c|}{Accuracy} & \multicolumn{2}{c}{Macro-F1} \\ \cmidrule{2-3}\cmidrule{4-5}
$\tau_s$ & LLaMA & Qwen & LLaMA & Qwen \\
\midrule
-1.0 & 76.0 & 76.9 & 60.8 & 58.9 \\
-0.8 & 76.0 & 75.2 & 61.1 & 59.2 \\
-0.6 & 75.0 & 76.1 & 59.7 & 61.6 \\
-0.4 & 75.6 & 76.4 & 60.6 & 62.4 \\
-0.2 & 76.9 & 76.4 & 62.7 & 62.4 \\
0.0  & 76.6 & 76.5 & 62.1 & 62.5 \\
0.2  & 76.9 & 76.7 & 62.4 & 63.0 \\
0.4  & 77.0 & 76.7 & 62.2 & 62.8 \\
0.6  & 77.1 & 76.6 & 62.2 & 62.7 \\
0.8  & 77.2 & 76.5 & 62.3 & 62.6 \\
1.0  & 77.2 & 76.5 & 62.3 & 62.6 \\
\bottomrule
\end{tabular}
\caption{Sensitivity to the stop-threshold $\tau_s$. Performance remains stable across a broad range of values for both backbones, with accuracy varying within a narrow band and macro-F1 showing only small fluctuation.}
\label{tab:tau_s}
\end{table}

\begin{table}[h]
\centering
\small
\begin{tabular}{c|cc|cc}
\toprule
& \multicolumn{2}{c|}{Accuracy} & \multicolumn{2}{c}{Macro-F1} \\ \cmidrule{2-3}\cmidrule{4-5}
$\tau_v$ & LLaMA & Qwen & LLaMA & Qwen \\
\midrule
0.0 & 76.4 & 76.3 & 61.5 & 61.8 \\
0.2 & 76.4 & 76.3 & 61.5 & 61.8 \\
0.4 & 76.4 & 76.3 & 61.5 & 61.8 \\
0.6 & 76.4 & 76.2 & 61.5 & 61.7 \\
0.8 & 76.6 & 76.2 & 62.1 & 61.7 \\
\bottomrule
\end{tabular}
\caption{Sensitivity to confidence threshold $\tau_v$. Results are highly stable across values.}
\label{tab:tau_v}
\end{table}

\section{Sensitivity to Retrieval Depth}
\label{app:retrieval}

We vary the number of retrieved passages on a 100-sample test subset and report the corresponding gold-evidence recall at each depth.

\begin{table}[h]
\centering
\small
\setlength{\tabcolsep}{3pt}
\begin{tabular}{c|c|ccccc}
\toprule
Top-$n$ & Recall (\%) & Acc. & F1$_{mc}$ & F1$_T$ & F1$_{HT}$ & F1$_F$ \\
\midrule
1  & 4.7  & 47.0 & 39.7 & 9.8  & 41.8 & 67.4 \\
3  & 11.5 & 57.0 & 54.0 & 40.9 & 47.2 & 73.8 \\
5  & 16.8 & 63.0 & 60.5 & 47.8 & 54.8 & 79.0 \\
10 & 25.8 & 63.0 & 59.8 & 47.6 & 53.5 & 78.2 \\
20 & 32.3 & 70.0 & 67.1 & 51.2 & 61.5 & 88.6 \\
\bottomrule
\end{tabular}
\caption{Sensitivity to retrieval depth and the corresponding gold-evidence recall.}
\label{tab:retrieval_sensitivity}
\end{table}

Overall performance generally improves as evidence recall increases, although the trend is not strictly monotonic. From Top-5 to Top-10, recall increases from 16.8\% to 25.8\%, while accuracy remains unchanged and macro-F1 decreases slightly, suggesting that additional passages may introduce distraction alongside greater evidence coverage. At Top-3 (11.5\% recall), RADAR achieves 54.0 macro-F1 and 47.2 \textsc{Half-True} F1, and at Top-20 (32.3\% recall), macro-F1 reaches 67.1.

\end{document}